\documentclass{article}

\usepackage[preprint]{spconfa4}
\usepackage{array}
\usepackage{booktabs}
\usepackage{cite}
\usepackage{graphicx}
\usepackage{amssymb,amsmath}
\usepackage{framed,multirow}
\usepackage{latexsym}
\usepackage{array}
\usepackage{balance}
\usepackage{booktabs}
\usepackage{caption}
\usepackage{color}
\usepackage{graphicx}
\usepackage{multirow}
\usepackage{times}
\usepackage{url}
\usepackage{threeparttable}
\usepackage{float}

\newcommand\etal{\emph{et~al.}}

\begin{document}


\title{Multi-channel Attentive Graph Convolutional Network With Sentiment Fusion For Multimodal Sentiment Analysis}
%

\name{Luwei Xiao$^{1,\dagger}$\thanks{$\dagger$ Equal contribution}, Xingjiao Wu$^{1,\dagger}$, Wen Wu$^{1,\star}$\thanks{$\star$ Corresponding author}, Jing Yang$^{1}$, Liang He$^{1,\star}$}
\address{\small{$^{1}$ School of Computer Science and Technology, East China Normal University, Shanghai, China}\\{\small \{louisshaw, Xingjiao.Wu\}@stu.ecnu.edu.cn, \{wwu\}@cc.ecnu.edu.cn, \{jyang, lhe\}@cs.ecnu.edu.cn}
}

\UseRawInputEncoding
\maketitle

\begin{abstract}
Nowadays, with the explosive growth of multimodal reviews on social media platforms, multimodal sentiment analysis has recently gained popularity because of its high relevance to these social media posts. Although most previous studies design various fusion frameworks for learning an interactive representation of multiple modalities, they fail to incorporate sentimental knowledge into inter-modality learning.  This paper proposes a Multi-channel Attentive Graph Convolutional Network (MAGCN), consisting of two main components: cross-modality interactive learning and sentimental feature fusion. For cross-modality interactive learning, we exploit the self-attention mechanism combined with densely connected graph convolutional networks to learn inter-modality dynamics. For sentimental feature fusion, we utilize multi-head self-attention to merge sentimental knowledge into inter-modality feature representations. Extensive experiments are conducted on three widely-used datasets. The experimental results demonstrate that the proposed model achieves competitive performance on accuracy and F1 scores compared to several state-of-the-art approaches.

\end{abstract}


\vspace{-0.2in}
\section{Introduction}
\vspace{-0.1in}
\label{sec:intro}

Sentiment analysis (SA) is one of the widely studied fields in the Natural Language Process (NLP), which aims to predict the individual's attitude towards a product or event \cite{zhou2020sk}. With the rapid development of Internet technology, the Multimodal Sentiment Analysis (MSA) increasingly attracts more attention in recent years \cite{wu2021survey}. The goal of MSA is to determine the sentiment of a video, an image or a text-based on multiple modal features (see Fig.1.)\cite{mai2021analyzing, kumar2020gated}. Understanding an individual's attitude towards a specific entity is helpful in a wide range of application domains \cite{baltruvsaitis2018multimodal} (e.g., business intelligence, recommendation system, government intelligence, etc.).

According to different fusion strategies, existing methods can be divided into the following two categories: a) Approaches that make an advance on the LSTM architecture \cite{zadeh2018memory, liang2018multimodal}, b) Frameworks that emphasize applying the expressiveness of tensors for multimodal representation \cite{zadeh2017tensor, liu2018efficient}.
\begin{figure}[t]
	
	\includegraphics[width=1.1\linewidth]{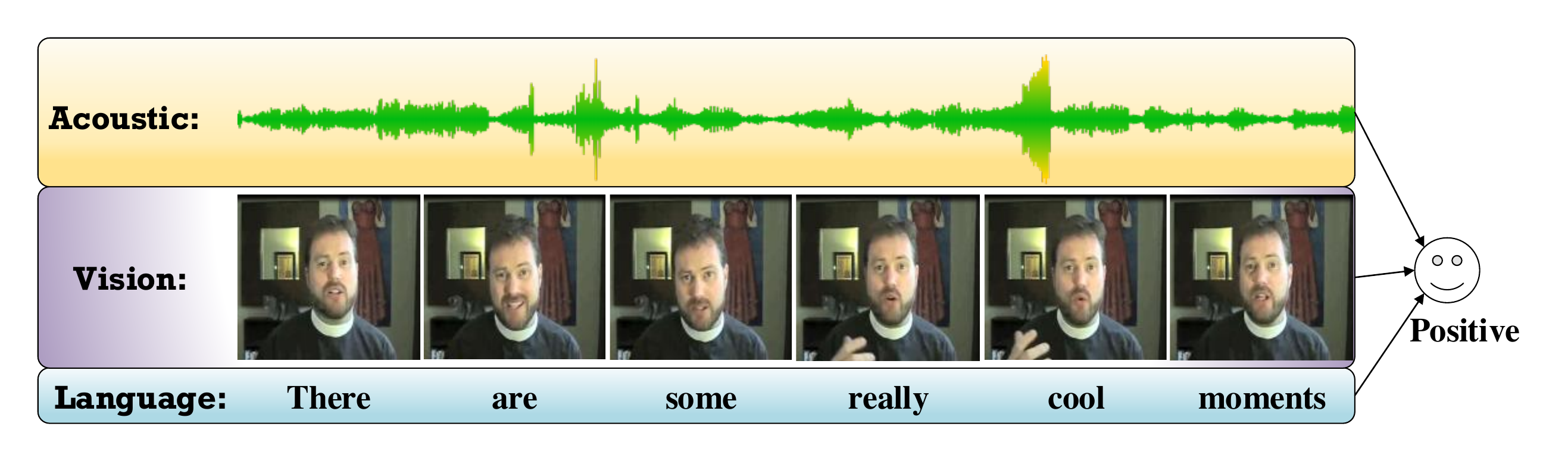}
	\caption{An example of multimodal sentiment analysis task. This task aims to predict the sentiment of a given utterance.
	}
	\label{fig_1}
	\vspace{-10px}
\end{figure}
Although previous studies have made salient improvements on benchmark datasets, there are still challenges for multimodal sentiment analysis. For one thing, most previous studies exploit simple operations (e.g., concatenation, addition) to learn inter-modality dynamics and conduct fusion within one step, which might weaken the overall performance of the multimodal framework since the cross-modal joint representation learned is less expressive. For another, few previous publications pay attention to incorporating the sentimental knowledge accompanying the language text into the multimodal system. Moreover, the importance of sentimental knowledge has been verified by sentence-level \cite{lei2018multi} and aspect-level \cite{tian2020skep, xiao2022exploring} sentiment analysis tasks, which can be adopted to assist the fusion of different modalities to be more sentiment-specific based on the rich sentimental information contained in the language. In order to solve the above problems, this paper proposes a novel Multi-channel Attentive Graph Convolutional Network (MAGCN) to fuse intra-/inter-modality with sentimental knowledge hierarchically. The main contribution of this paper is as follows:
\vspace{-0.1in}
\begin{itemize}
	\item We introduce sentiment embedding to make the use of sentimental knowledge for learning sentiment-specific inter-/intra-modality fusion representation.
	\vspace{-0.15in}
	\item We introduce densely connected graph convolutional network from relation extraction task to multimodal sentiment analysis, which successfully models the inter-/intra-modality feature representations.
	\vspace{-0.15in}
	\item  We define a novel consistency constraint loss to further enhance the commonality between modalities.
	\vspace{-0.15in}
	\item The proposed model yields new state-of-the-art results on three MSA benchmark datasets.
	\vspace{-0.15in}
\end{itemize}
\vspace{-0.2in}
\section{Related Work}
\vspace{-0.1in}
For approaches that make an advance on the LSTM architecture, Zadeh \etal~\cite{zadeh2018memory} designed a novel Memory Fusion Network which is composed of a System of LSTMs and Delta-memory Attention Network to learn both the view-specific interactive representations and the cross-view interactive representations. Liang \etal~\cite{liang2018multimodal} put forward a Recurrent Multistage Fusion Network to learn cross-modal interactive information by exploiting a specialized multistage fusion approach. For methods that lay emphasis on applying the expressiveness of tensors for multimodal representation, Zadeh \etal~\cite{zadeh2017tensor} proposed a Tensor Fusion Network to explicitly model the intra-modality and inter-modality dynamics end-to-end via a 3-fold Cartesian product from the modality embedding. Liu \etal~\cite{liu2018efficient} exploited a Low-rank Multimodal Fusion network that first obtains the unimodal representation and leverage low-rank weight tensors to finish multimodal fusion.

Recently, some other approaches have achieved new state-of-the-art results. Pham \etal~\cite{pham2019found} designed a framework of learning joint representations, a cycle consistency loss is proposed to ensure that the joint representations retain maximal information from all modalities. Mai \etal~\cite{mai2019divide} devised a ?divide, conquer and combine? strategy to fuse cross-modality dynamically and hierarchically. Tsai \etal~\cite{tsai2018learning} utilized a Multimodal Factorization Model which factorizes features into multimodal discriminative and modality-specific generative factors and designs a common generative-discriminative objective to optimize across multimodal data and labels. Chen \etal~\cite{chen2020swafn} introduced a method consists of a shallow fusion and aggregation module, and an auxiliary sentimental words classification is added to assist and guide the profound fusion of the three modalities.

\vspace{-0.2in}
\section{APPROACH}
\vspace{-0.1in}
\label{sec:format}

The overall architecture of MAGCN is illustrated in Fig.2. This section will describe our approach in more detail, including the contextual utterance representation, the inter-modality representation learning, the unimodal language representation learning, and the inference module.
\vspace{-0.1in}

\begin{figure}[t]	
	\includegraphics[width=1.0\linewidth]{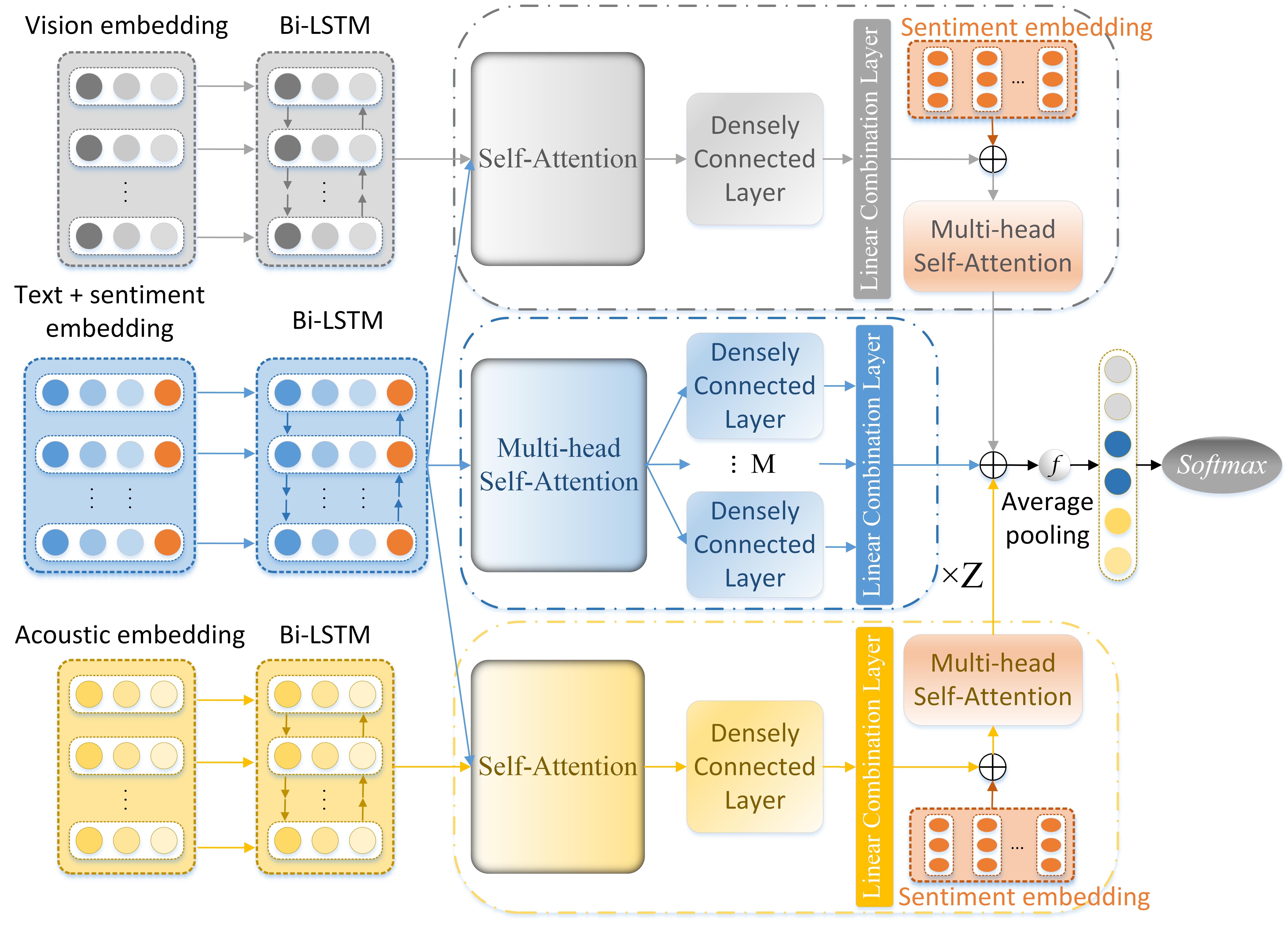}
	\caption{The overall architecture of proposed MAGCN.
	}
	\label{fig_2}
	\vspace{-8px}
\end{figure}

\subsection{Contextual Utterance Representation}
\vspace{-0.1in}
The sentiment knowledge in each word should be incorporated into the learning process of intra-modality and inter-modality to generate richer multimodal representation. Inspired by Chen \etal~\cite{chen2020swafn}, we propose to utilize Bing Liu's Opinion Lexicon \cite{ding2008holistic} to train a sentiment embedding for each word as the external knowledge. The function of sentiment embedding is to indicate whether a word is a sentimental word (If a word is in the Opinion Lexicon, then it is). The weight of each dimension for sentiment embedding is trainable during the whole process. $S = \left\{ {{s_1},{s_2}, \cdots ,{s_n}} \right\}$ refers to sentiment embedding for input sequence. Given the original representation of text ${E} = \left\{ {{e_1},{e_2}, \cdots ,{e_n}} \right\}$, acoustic ${X_A} = \left\{ {{a_1},{a_2}, \cdots ,{a_n}} \right\}$ and vision ${X_V} = \left\{ {{v_1},{v_2}, \cdots ,{v_n}} \right\}$, we concatenate the text embedding and sentiment embedding. The representation for a word is ${x_i} = \left[ {{e_i};{s_i}} \right]$, the language modality is denoted as ${X_L} = \left\{ {{x_1},{x_2}, \cdots ,{x_n}} \right\}$. We first exploit the Bi-LSTM to model the representation of the above three modalities and obtain the feature encoding of them, denoted as ${H_L}$, ${H_A}$ and ${H_V}$, respectively.

\vspace{-0.1in}
\subsection{Inter-Modality Representation Learning}
\vspace{-0.1in}
In this part, we combine self-attention, densely connected graph convolutional networks with multi-head self-attention to learn the cross-modal feature representation between language modality and other modalities. The self-attention mechanism \cite{cheng2016long} and densely connected graph convolutional networks (DCGCN) \cite{guo2019attention} are exploited to learn the bimodal fusion between language modality and other modalities (i.e., vision or acoustics) simultaneously. Firstly, an affinity matrix is generated by self-attention mechanism:
\vspace{-0.1in}
\begin{equation}
{A^{LV}} = soft\max \left( {\frac{{H_L{W_Q} \times {{\left( {{H_V}{W_K}} \right)}^T}}}{{\sqrt d }}} \right)
\vspace{-0.1in}
\end{equation}
which can be treated as a fully connected edge-weighted graph. These weights can be viewed as the strength of relatedness between language modality and vision (or acoustics) modality. ``$d$" denotes the dimension of ${H_L}$. ${W_Q} \in {R^{d \times d}}$ and ${W_K} \in {R^{d \times d}}$ are both parameter matrices.
Next, the DCGCN \cite{guo2019attention} is introduced to capture rich local and non-local dependency information between different modalities. $g_j^l = \left[ {{x_j};h_j^1; \cdots ;h_j^{l - 1}} \right]$ is denoted as the concatenation of original node feature embedding and other nodes embedding generated in layers $1,2, \cdots ,l - 1$, where ${{x_j}}$ refers to the initial input for DCGCN, produced from ${H_V}$ (or ${H_A}$). In DCGCN, each densely connected layer has $L$ sublayers. The dimension of each sublayer is determined by the input dimension $d$ and $L$. For instance, if the input dimension is 60 and $L=3$, the hidden dimension of each sub-layer will be ${d_{sub}} = d/L = 60/3 = 20$. An example of DCGCN is shown in Fig.3. Each layer of DCGCN is computed as follows:
\vspace{-0.1in}
\begin{equation}
h_i^l = \sigma (\sum\limits_{j = 1}^n {A_{ij}^{LV}W_{}^lg_j^l}  + b_{}^l)
\vspace{-0.1in}
\end{equation}
where $W_{}^l \in {R^{{d_{sub}} \times {d^l}}}$ and $b_{}^l \in {R^{{d_{sub}}}}$ are trainable matrix and bias, respectively. ${d^l} = d + {d_{sub}} \times \left( {l - 1} \right)$. After we feed the fully connected edge-weighted graph $A^{LV}$ (or $A^{LA}$) and ${H_V}$ (or ${H_A}$) into DCGCN, we obtain the bimodal fusion representation ${H_{LV}}$ (or ${H_{LA}}$). Then, a linear transformation is followed to obatin the output of DCGCN $H_{LV}^{gcn}$. Finally, we concatenate the bimodal fusion representation ${H_{LV}}$ (or ${H_{LA}}$) with sentiment embedding $S$ as the input $H{S_{LV}}$ (or $H{S_{LA}}$) for multi-head attention, learning the sentiment-awareness bimodal representation $HS_{LV}^{out}$ (or $HS_{LA}^{out}$) :
\vspace{-0.1in}
\begin{equation}
HS_{LV}^{out} = MHA\left( {H{S_{LV}},H{S_{LV}}} \right)
\end{equation}
where $MHA$ is multi-head self-attention \cite{vaswani2017attention}. $HS_{LV}^{out}$ is the final inter-modality feature representation of the language modality and visual modality. The final inter-modality feature representation of the language modality and acoustic modality $HS_{LA}^{out}$ can be generated in the same way.

\begin{figure}[t]
	
	\includegraphics[scale=0.7]{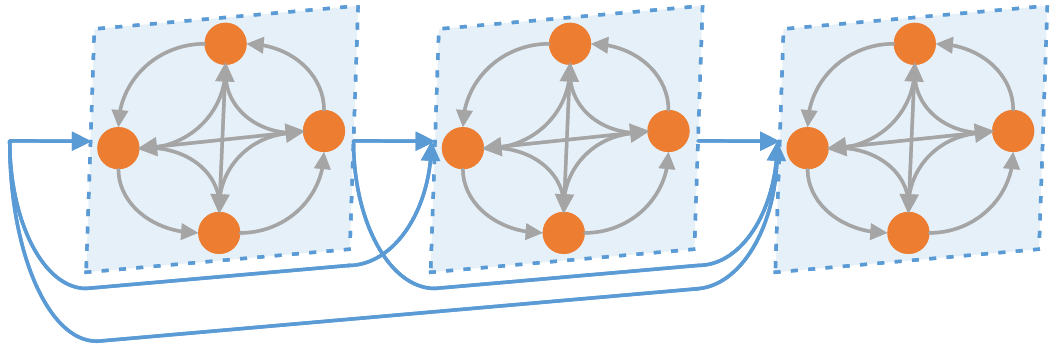}
	\caption{The architecture of densely connected graph convolution networks(DCGCN).
	}
	\label{fig_3}
	\vspace{-8px}
\end{figure}

\vspace{-0.1in}
\subsection{Unimodal Language Representation Learning}
\vspace{-0.1in}
This module comprises $Z$ identical blocks. Each block is composed of multi-head self-attention and DCGCN for extracting richer semantic information from context representations. Given the language modality representation ${H_L}$, we first generate several fully connected graphs by exploiting multi-head self-attention:
\vspace{-0.1in}
\begin{equation}
{\widetilde B^{(t)}} = {\mathop{\rm softmax}\nolimits} \left( {\frac{{{H_L}{\bf{W}}_t^Q \times {{\left( {{H_L}{\bf{W}}_t^K} \right)}^T}}}{{\sqrt {d/M} }}} \right)
\end{equation}
where ${\widetilde B^{(t)}} \in {R^{n \times n}}$ is the $t$-th matrix constructed corresponding to the $t$-th head. Up to $M$ matrices are constructed in total. For $M$ attention graphs, $M$ independent DCGCN are required. Each layer of DCGCN is computed as follows:
\vspace{-0.1in}
\begin{equation}
h_{{t_i}}^l = \sigma (\sum\limits_{j = 1}^n {\widetilde B_{ij}^{(t)}W_t^lu_j^l}  + b_t^l)
\end{equation}
where $u_j^l = \left[ {h_j^0;h_j^1; \ldots ;h_j^{l - 1}} \right]$. $h_j^0$ is obtained from ${H_L}$. $t \in \left[ {1,2, \ldots ,M} \right]$. With different matrices, different parameter matrix $W_t^l \in {R^{{d_{sub}} \times {d^l}}}$ and bias term $b_t^l \in {R^{{d_{sub}}}}$ are picked up for training. Then, the context feature representation from $M$ different DCGCN are concatenated, a linear transformation is exploited to obtain the final feature representation of language modality:
\vspace{-0.1in}
\begin{equation}
H_L^{out} = {W_{out}}\left[ {{h^1};{h^2}; \cdots ;{h^M}} \right] + {b_{out}}
\end{equation}
where ${W_{out}} \in {R^{\left( {d \times M} \right) \times d}}$ and ${b_{out}} \in {R^d}$ are trainable matrix and bias, respectively.
\vspace{-0.1in}
\subsection{Decision Inference Module}
\vspace{-0.1in}
In order to improve the commonality between the inter-modality representation, we design a consistency loss function for achieving this goal. ${L_2}$-normalization is used to normalize the embedding matrices $H_L^{out}$,  $HS_{LV}^{out}$ and $HS_{LA}^{out}$ and denoted them as $C_{Lnor}$, $C_{LVnor}$ and  $C_{LAnor}$, respectively. Then, they can be used to guide the inter-modality representations to enhance their commonality further as follows:
\vspace{-0.1in}
\begin{equation}
{L_c} = \left\| {{C_{Lnor}} \cdot C_{LVnor}^T - {C_{Lnor}} \cdot C_{LAnor}^T} \right\|_F^2
\vspace{-0.1in}
\end{equation}

Finally, we concatenate $H_L^{out}$, $HS_{LV}^{out}$ and $HS_{LA}^{out}$, and an average pooling function is applied, the final feature representation for our model is obtained, which is then feed into $softmax$ to produce the sentiment prediction ${y_s}$. The overall loss function of our model is shown as follows:
\vspace{-0.1in}
\begin{equation}
Loss = \alpha \left( {\frac{1}{N}\sum\limits_{i = 1}^N {\left| {y_s^i - \hat y_s^i} \right|} } \right) + \beta {L_c}
\vspace{-0.1in}
\end{equation}
where $\alpha$ and $\beta$ are hyper-parameters. $N$ denotes all training samples in the dataset. $y_s^i$ and $\hat y_s^i$ are predicted sentiment and true sentiment for $i$-th sample, respectively.

\vspace{-0.2in}
\section{Experiments}
\vspace{-0.2in}
\subsection{Datasets}
\vspace{-0.1in}
CMU-MOSI \cite{zadeh2016mosi},CMU-MOSEI \cite{zadeh2018multimodal} and YouTube \cite{morency2011towards} are selected as our experimental datasets. GloVe \cite{pennington2014glove} is utilized to generate the language embedding, the visual representations are obtained by Facet library\footnote{https://imotions.com/biosensor/fea-facial-expression-analysis/} and acoustic features are extracted by applying COVAREP \cite{degottex2014covarep}.

\vspace{-0.2in}
\subsection{Baseline Models}
\vspace{-0.1in}
Some classical studies \cite{zadeh2017tensor, liu2018efficient} and recently state-of-the-art approaches \cite{pham2019found, mai2019divide, chen2020swafn, tsai2018learning}  which are introduced in Section 2 have been chosen as baseline models for experiments. Accuracy and F1 scores are picked up as our evaluation metrics.
\vspace{-0.2in}
\subsection{Experimental Results and Analysis}
\vspace{-0.1in}
The experimental results on YouTube, MOSEI and MOSI datasets are shown in Table~\ref{tab_1}. From Table~\ref{tab_1}, there are a certain number of issues worth noting. Firstly, For the YouTube dataset, compared with the best baseline model SWAFN, the MAGCN gives a higher overall performance, which outperforms SWAFN by 1.2$\%$ on accuracy and 1.3 $\%$ on F1 score. Since the YouTube dataset contains relatively few samples, most baselines may be overfitting on its training set. However, our model achieves the best performance, demonstrating the better generalization capability of the proposed model. Secondly, for MOSEI and MOSI datasets, MAGCN still achieves the best performance over all evaluations. One characteristic these baselines have in common is that they treat language modality equally with other modalities. In contrast, we specifically implement a language feature learning module in a more straightforward and powerful way and design a consistency loss function to partially guide other modalities to focus on features from language modality. All in all, the experimental results on three datasets demonstrate the effectiveness and robustness of the proposed model.

\begin{table}[t]
	\caption{Experimental results ($\%$). ``N/A" refers to unreported results. Highest results are in bold.}
	\begin{tabular}{lcccccc}
		\toprule
		\multicolumn{1}{c}{\multirow{2}{*}{Model}} &
		\multicolumn{2}{c}{YouTube} &
		\multicolumn{2}{c}{CMU-MOSEI} &
		\multicolumn{2}{c}{CMU-MOSI} \\  \cline{2-7}
		\multicolumn{1}{c}{} & Acc  & F1   & Acc   & F1    & Acc  & F1   \\ \midrule	
		LMF \cite{liu2018efficient} & N/A  & N/A  & 60.27 & 53.87 & 76.4 & 75.7 \\
		TFN \cite{zadeh2017tensor}  & 45.0 & 41.0 & 59.4  & 57.33 & 74.6 & 74.5 \\
		MFM \cite{tsai2018learning} & 53.3 & 52.4 & N/A   & N/A   & 78.1 & 78.1 \\
		MCTN \cite{pham2019found}   & 51.7 & 52.4 & N/A   & N/A   & 79.3 & 79.1 \\
		HFFN \cite{mai2019divide}   & N/A  & N/A  & 60.37 & 59.07 & 80.2 & 80.3 \\
		SWAFN \cite{chen2020swafn}  & 55.0 & 53.3 & 61.03 & 59.32 & 80.2 & 80.1 \\
		MAGCN &  \textbf{56.2} &\textbf{54.6} &\textbf{61.95} &\textbf{59.70} &\textbf{80.6} &\textbf{80.3} \\\bottomrule
	\end{tabular}
	\vspace{-8px}
	\label{tab_1}
\end{table}

\vspace{-0.15in}
\subsection{Ablation Study}
\vspace{-0.1in}
\noindent
\textbf{Contribution of Different Modalities}
We conduct a series of different experiments on the MOSI dataset to investigate the influence of each modality. Table 2 displays the experimental results of using unimodal, bimodal and multimodal features. Firstly, from the unimodal perspective, the use of language features outperforms acoustic features or vision features significantly, which verifies that language modality contains much more crucial information than other modalities. Secondly, from a bimodal perspective, we notice that although the combination of language features and acoustic features or vision features improves the accuracy compared with only utilizing language features, not all F1 scores can be improved. Besides combining acoustic features with video features, the model still gives frustrating results, indicating that the language modality should be dominant for this task. Thirdly, for incorporating three modalities, with the assistance of sentimental knowledge and consistency loss as well as DCGCN, MAGCN achieves state-of-the-art performance.

\begin{table}[]
	\caption{Experimental results ($\%$) of proposed model using unimodal, bimodal and multimodal features. ``V" means Video, ``A" means Audio, ``L" means Language.}
	\label{tab_2}
	\begin{tabular}{@{}lccccccc@{}}
		\toprule
		\multicolumn{1}{c}{} & \multicolumn{3}{l}{Unimodal} & \multicolumn{3}{l}{Bimodal} & Multimodal \\ \hline
		& V        & A       & L       & A+V      & L+V     & L+A     & L+V+A      \\ \midrule
		Acc                  & 59.1     & 58.6    & 78.5    & 59.4     & 78.8    & 79.1    & \textbf{80.6}       \\
		F1                   & 59.2     & 58.4    & 78.7    & 59.3     & 78.5    & 78.9    & \textbf{80.3}       \\ \bottomrule
	\end{tabular}
	\vspace{-8px}
\end{table}

\vspace{0.06in}
\noindent
\textbf{Contributions of Sentiment Embedding and Consistency Loss}
To evaluate the contributions of sentiment embedding and consistency loss and DCGCN, ablation studies are performed. In Table~\ref{tab_3}, the removal of sentiment embedding leads to a drop in both the accuracy and F1 score over two datasets, which indicates that sentiment knowledge is beneficial to this task. We find that for MOSEI, the removal of sentiment knowledge leads to worse performance than the lack of consistency loss. In contrast to the MOSEI, the MOSI dataset tends to highlight the information from language modality since removing the consistency loss leads to the dramatic fall of about 2$\%$ in accuracy and F1 score. We suppose that samples from MOSI are more context-aware while data in MOSEI contain more useful sentimental information. The results demonstrate the effectiveness of consistency loss in guiding the acoustic and vision modality to learn more details from language representation, highlighting the necessity of carefully analyzing human language in MSA.

\noindent
\textbf{Contributions of Densely Connected Graph Convolutional Neural Nwtworks}  In Table 3, we also use vanilla GCN to replace DCGCN. Without dense connection, the model is deficient in accurately capturing latent semantic correlations between nodes, especially those connected by in-direct, multi-hop paths cross-modality relations, which causes disappointing performance on both datasets. The results demonstrate that DCGCN is quite crucial for our framework.

\begin{table}[t]
	\caption{Ablation study results ($\%$). Highest results are in bold. ``w/o SE" means no sentiment embedding is exploited. ``w/o CL" means no consistency loss is applied. ``w/o DCGCN" indicates that DCGCN is replaced by Vanilla GCN.}
	\begin{tabular}{llllll}
		\toprule
		\multicolumn{2}{l}{\multirow{2}{*}{Model}} & \multicolumn{2}{l}{CMU-MOSEI}   & \multicolumn{2}{l}{CMU-MOSI}    \\ \cline{3-6}
		\multicolumn{2}{l}{}                       & Acc            & F1             & Acc            & F1             \\ \hline
		\multicolumn{2}{l}{MAGCN}                  & \textbf{61.94} & \textbf{59.70} & \textbf{80.61} & \textbf{80.32} \\ \hline
		\multicolumn{2}{l}{MAGCN w/o SE}           & 60.88          & 58.41          & 79.43          & 79.13          \\
		\multicolumn{2}{l}{MAGCN w/o CL}           & 61.29          & 59.25          & 78.82          & 78.61          \\
		\multicolumn{2}{l}{MAGCN w/o DCGCN}        & 61.08          & 59.15          & 79.27          & 79.05          \\ \bottomrule
	\end{tabular}
	\label{tab_3}
	\vspace{-8px}
\end{table}

\vspace{-0.2in}
\section{Conclusion}
\vspace{-0.1in}
In this paper, we propose a framework to improve the MSA by using self-attention combined with DCGCN to learn inter-modality feature representation, then fuse sentiment knowledge via multi-head self-attention and exploiting a module consists of multi-head self-attention and DCGCN to extract intra-language feature representation. A consistency loss is designed to enhance the commonality between inter-modality feature representations. Experiments on three well-known datasets (YouTube, CMU-MOSI and CMU-MOSEI) show the effectiveness of the proposed model.

\vspace{-0.2in}
\section{Acknowledgements}
\vspace{-0.1in}
This work was funded by grants from the National Natural Science Foundation of China (under project No. 61907016) and the Science and Technology Commission of Shanghai Municipality (under project No. 21511100302).

\small{
\bibliographystyle{IEEEtran}
\bibliography{egbib}}

\end{document}